\documentclass[conference, uspaper]{IEEEtran}

\usepackage{multirow}
\usepackage[hidelinks]{hyperref}
\usepackage{booktabs}
\usepackage{cite}
\usepackage[pdftex]{graphicx}
\graphicspath{{..figures/}{../figures/}}
\DeclareGraphicsExtensions{.pdf,.jpeg,.png}
\usepackage{amsmath}
\usepackage{siunitx}

\begin{document}

\title{Physical LiDAR Simulation in Real-Time Engine}

 \author{\IEEEauthorblockN{Wouter Jansen\IEEEauthorrefmark{1}\IEEEauthorrefmark{2}\IEEEauthorrefmark{3}, Nico Huebel\IEEEauthorrefmark{1}\IEEEauthorrefmark{2}, Jan Steckel\IEEEauthorrefmark{1}\IEEEauthorrefmark{2}}
 \IEEEauthorblockA{\IEEEauthorrefmark{1}FTI Cosys-Lab, University of Antwerp, Antwerp, Belgium}
 \IEEEauthorblockA{\IEEEauthorrefmark{2}Flanders Make Strategic Research Centre, Lommel, Belgium\\
 \IEEEauthorrefmark{3}wouter.jansen@uantwerpen.be}}
 
\maketitle

\begin{abstract}
Designing and validating sensor applications and algorithms in simulation is an important step in the modern development process. Furthermore, modern open-source multi-sensor simulation frameworks are moving towards the usage of video-game engines such as the Unreal Engine. Simulation of a sensor such as a LiDAR can prove to be difficult in such real-time software. In this paper we present a GPU-accelerated simulation of LiDAR based on its physical properties and interaction with the environment. We provide a generation of the depth and intensity data based on the properties of the sensor as well as the surface material and incidence angle at which the light beams hit the surface. It is validated against a real LiDAR sensor and shown to be accurate and precise although highly depended on the spectral data used for the material properties. \\
\end{abstract}


\IEEEpeerreviewmaketitle

\section{Introduction}
\label{sec:introduction}
There are many existing simulation frameworks in robotics, both closed\cite{siemensprescan,ansys,dyna4,vtd,driveworks,rfpro,aurelion} and open-source\cite{Shah2017AirSim:Vehicles,Rong2020LGSVLDriving,webots,Dosovitskiy2017CARLA:Simulator}. Techniques such as real-time rendering, particularly with video-game engines such as Unity and Unreal have caused accelerated development of open-source simulation platforms used by many organisations and researchers for experimental development, research, validation and comparison against other systems and algorithms. One vital aspect of any simulation platform is to provide accurate sensor models without unrealistic or too idealistic assumptions while balancing the complexity and subsequent computational demands of such sensor models against the simulation performance.\\
Within this paper we present our implementation of the simulation of Light Detection And Ranging (LiDAR) sensors in an open-source framework using the real-time Unreal Engine. These provide a 3D depth map of the environment (in the form of a point cloud). However, to our knowledge most documented and available simulations have one or multiple shortcomings: (a) they are often build for addressing a specific target problem and do not come in a modular framework together with other sensors \cite{Gusmao2020DevelopmentRaycasting, Manivasagam2020LiDARsim:World, Fang2018AugmentedDriving, Hanke2018GenerationSystems, Wang2009, Vasstein2020AutoferryShips}; (b) they provide a perfect rendition of the perceived environment with simple noise models\cite{Shah2017AirSim:Vehicles, Rong2020LGSVLDriving, Dosovitskiy2017CARLA:Simulator}, or (c) they don't allow for realistic environments to use LiDAR sensors\cite{Koenig2004DesignSimulator, mathworksautomation}. LiDAR is projecting beams of structured light at a specific wavelength with a certain power, which is subject to distortions from environmental factors such as the properties of the reflecting surfaces. The surface properties are mainly defined by the angle between the light beam and the surface as well as the material of the surface, which both affect the reflection intensity perceived by the LiDAR \cite{Muckenhuber2020AutomotiveCapabilities,Nicodemus1965DirectionalSurface}.\\
The major contribution of this research is a realistic real-time simulation of a high resolution 3D LiDAR sensor in complex environments by including angle dependant material reflectance properties. Furthermore, other properties of the LiDAR sensor such as its beam configuration and range capabilities are also modelled and implemented. The implemented sensor models were based on the existing work of other peer-reviewed publications. To ensure the real-time performance, some the computations are done on the GPU. Instance semantic ground-truth labels are also provided for each point. We validate the simulation by comparing real-world measurements against a re-created, simulated environment and compare the accuracy and precision of the resulting reflectance and depth of each point for various material types and different ranges between the objects and sensor. Finally, we provide an open-source release\cite{Cosys-LabGPULiDAR-AirSimRepository} of our LiDAR simulation for the AirSim framework \cite{Shah2017AirSim:Vehicles} so it can be cross-validated and used by others. 

\begin{figure}[ht]
\centering
\includegraphics[width=0.96\linewidth]{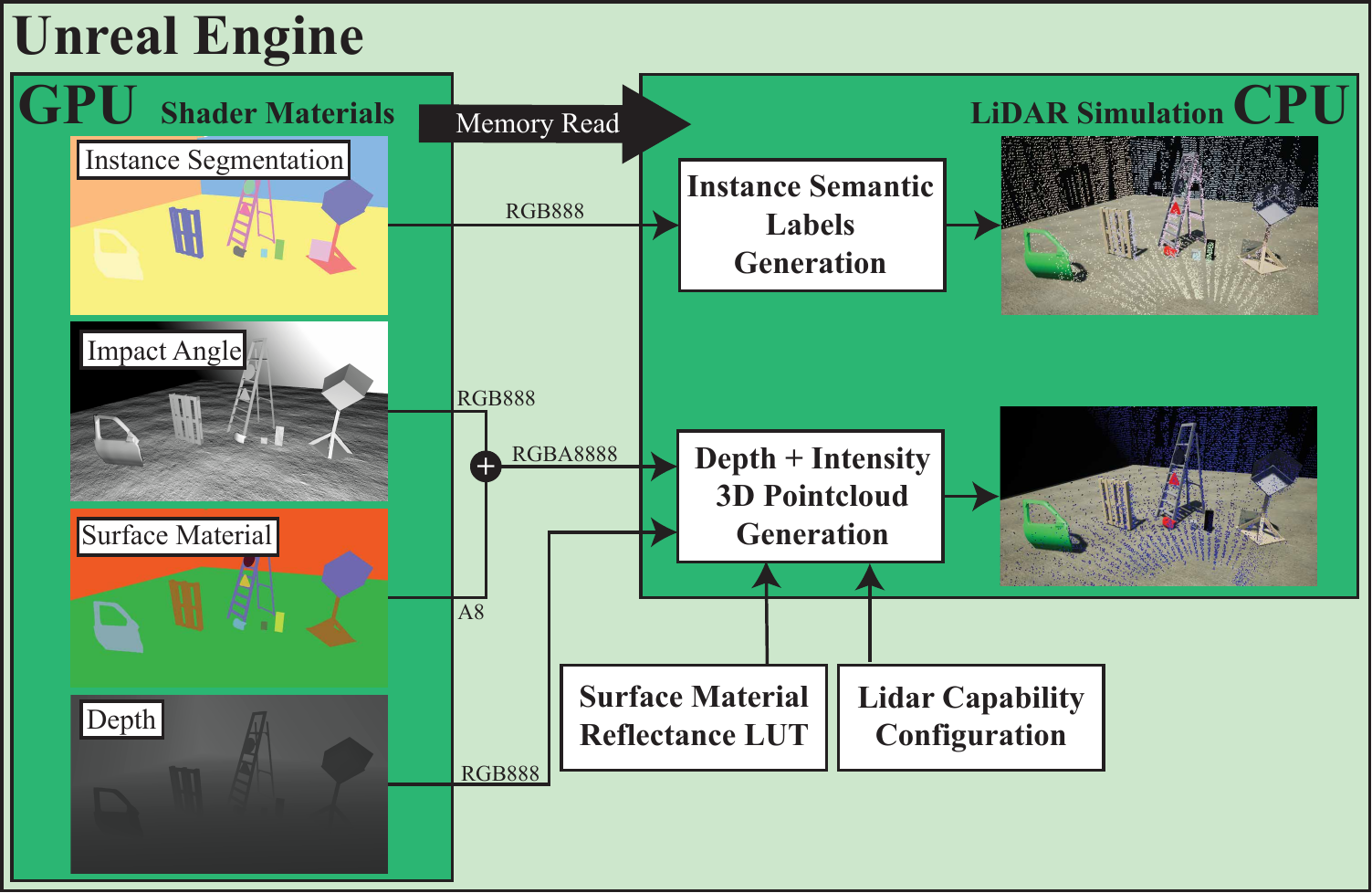}
\caption{Overview of the GPU-accelerated LiDAR simulation in Unreal Engine. Several shader materials are generated on the GPU which are subsequently read on the CPU from the textures.
Their output is used together with a look-up-table of surface spectral data and the sensor configuration to generate a 3D point cloud with the depth, intensity and instance segmentation label data.}
\label{fig:overview}
\end{figure}

\section{Simulation Implementation}
\label{sec:implementation}
Our method for simulating LiDAR sensors is implemented within our version of the AirSim framework\cite{Shah2017AirSim:Vehicles,Schouten2021SimulationSLAM} as it provides a wide range of mobile platforms and a stable API for control and sensor data extraction. As AirSim is built upon the Unreal Engine, its real-time render pipeline is used to generate the LiDAR point cloud data. Each subsection will provide a more detailed explanation on how the data is generated. An overview of the implementation is shown in Fig. \ref{fig:overview}.

\subsection{Depth Point Cloud Generation}
\label{subsec:pointcloudgeneration}
The base of any LiDAR simulation is the depth data generation. Our method uses a virtual render camera that for each time step of the simulation will be rotated according to the elapsed as the simulation is real-time and not fixed step. This camera has a single post-process material shader that uses the depth pass of the engine G-Buffer\cite{EpicGamesUnrealOverview}. This serves as the base of the GPU-accelerated LiDAR simulation method we use. In order to pertain the highest precision possible, the depth floating point value of the G-Buffer is saved by the shader over the RGB channels of the material to achieve three bytes unsigned precision. Subsequently, this render camera is parsed and read out by the CPU and stored in a CPU memory buffer. This camera has a fixed configurable square resolution of width $W$.\\
The virtual camera uses the pin-hole model and has a dynamic Field Of View (FOV) that is calculated before rendering at each time step by knowing how much of the LiDAR's horizontal FOV $FOV_h$ needs to be parsed based on the sensor's spinning frequency $\omega_l$ and the time delta to the previous time step. As the resolution is square, the vertical and horizontal FOV of the virtual camera are equal and defined as $FOV_c$. With the camera buffer now available on the CPU, the next step is to sample this image for each of the LiDAR's beams based on the horizontal and vertical sample points that are predefined. In order to do so we need to convert from the spherical coordinates of the LiDAR beams to the pixel-based image space coordinates. The virtual render camera's intrinsic matrix $K_c$ can be defined as that of an ideal camera without distortion:
\begin{equation}
K_c = 
\begin{bmatrix}
    f_x & 0 & c_u  \\
    0 & f_y & c_v  \\
    0 & 0 & 1
\end{bmatrix}
\label{eq:camera_proj_matrix}
\end{equation}

with the focal lengths $f_x$ and $f_y$ and principal camera point $(c_u,c_v)$ defined as
\begin{equation}
f_x = f_y = \frac{W}{2\tan{\cfrac{FOV_c}{2}}}\quad c_u = c_v = \frac{W}{2}
\label{eq:focal_and_principal_points}
\end{equation}

Subsequently, for a single light beam $l_{i,j}$ of the LiDAR with horizontal sample index $i$ and vertical index $j$ and spherical coordinates $(\theta_i,\phi_j)$ based on the LiDAR properties $FOV_v$ and $FOV_h$ for their vertical and horizontal FOVs respectively, we can calculate the pixel coordinates $(n_x^i,n_y^j)$.
\begin{equation}
n_x^i = \cfrac{(\sin{\theta_i}f_x)}{\cos{\theta_i}} + c_u \quad
n_y^j = \cfrac{(\sin{\phi_j} -f_y)}{\cos{\phi_j} \cos{\theta_i}} + c_y
\label{eq:pixel_equations}
\end{equation}

The azimuth $\theta_i$ is defined by the horizontal FOV $FOV_h$ of the sensor, which is usually $[\ang{0},\ang{360}]$ for a \ang{360} LiDAR sensor. Similarly, The elevation $\phi_j$ is defined by the vertical FOV $FOV_v$ and is defined in the sensor data sheets how the different beams are spread out.\\ Then the depth value for this light beam can be retrieved from the CPU buffer at these coordinates. Gaussian noise is also added within this model as a separate property to define a noise scaling factor. 

\subsection{Instance Semantic Ground-truth}
\label{subsec:instancesemanticgt}
While AirSim already includes ground-truth labels for objects classes, it is limited to 255 different classes as it relies on the custom depth buffer (a single byte value) of the Unreal renderer. UnrealCV\cite{QiuUnrealCV:Engine,Qiu2017UnrealCV:Vision} provides a solution to uniquely label millions of object in the virtual world by overwriting their vertex colours with a unique RGB88 colour that is saved in a data table. We applied this solution to the LiDAR sensor within the AirSim framework and upgraded it to the latest versions of the Unreal Engine. Similar to the depth data generation, a virtual render camera was used, but with an additional post processing material shader to only show the vertex colours. Using the sampling technique described for the point cloud generation, the RGB888 colour value for each LiDAR point can be obtained. This value allows to find the object name and class of millions of unique objects from the automatically generated data table.

\subsection{Surface Material Model}
\label{subsec:surfacematerialmodel}
LiDAR is influenced by material properties of the objects it observes. The used model is based on the work by Muchkenhuber et al. \cite{Muckenhuber2020AutomotiveCapabilities}. We refer to their article for extensive details. We integrate their model for Lambertian target reflectance $R_{\lambda}$, a value in percentage describing the reflectance measured for wavelength $\lambda$ at various incidence angles $\theta$. Their work refers to the ECOSTRESS spectral library \cite{JetPropulsionLaboratoryECOSTRESS1.0} which includes the spectra of thousands of materials measured at various wavelengths including some that are commonly seen in LiDAR applications. The samples in this library represent the reflectance in percentage for a incidence angle $\theta$ of 0° $R_{\lambda}(\theta = 0^{\circ})$.
To include the influence of the incidence angle of the LiDAR beam in the material model, we can derive the Lambertian target reflectance $R_{\lambda}$ for a certain material with a specific incidence angle as:
\begin{equation}
R_{\lambda}(\theta) = R_{\lambda}(\theta = 0^{\circ})  \cos(\theta)
\label{eq:material}
\end{equation}

The incidence angle of the LiDAR can be retrieved in the Unreal Engine similarly to the other implementations using a custom material shader. In order to calculate the incidence angle of the LiDAR light beam to each point in the environment that can be seen from the virtual render camera, we make use of the dot product of the surface normal, which is the unit vector perpendicular to the surface, and the vector representing the LiDAR light beam. 
To achieve three bytes precision, this angular value is stored in the shader in a RGBA8888 texture. \\ 
This leaves the surface material to be acquired for each point. We chose to use the stencil buffer of Unreal, a custom depth buffer of a single byte value per pixel. This limits the use to 255 different materials. We attribute each integer value to a unique material of the ECOSTRESS spectral library for the wavelength $\lambda$ in use by the LiDAR, which is stored at rendering in the alpha channel of the previously mentioned RGBA8888 output texture. By parsing this texture on the CPU we can extract the material and the incidence angle and use (\ref{eq:material}) to calculate their influence on the reflectance and intensity.

\subsection{LiDAR Capability Model}
\label{subsec:propertymodel}
The optical performance of a LiDAR sensor is defined in its data sheets. This includes the reflectance limit function $R_L(d)$, which depicts the maximum range $d$ for which a point would be detected depending on the corresponding Lambertian target reflectance $R_{\lambda}$. 
For our LiDAR capability model we implemented a linear function $R_L(d)$ with a configurable maximum range $d_{max}$ and the corresponding target reflectance $R_{L,max}$, which is needed to detect a point at that maximum range. Then, between a range of \SI{0}{\m} and $d_{max}$ the function interpolates $R_{\lambda}$ linearly. Any point beyond $d_{max}$ is dropped.
\begin{equation}
R_L(d) = \cfrac{R_{L,max}}{d_{max}} \times d
\label{eq:capability}
\end{equation}

For any point its depth $d$ (including Gaussian noise) and reflectance intensity $R_{\lambda}$ from model described in Section \ref{subsec:surfacematerialmodel} are obtained. Then this linear function is used to decide if the point is detected ($R_L(d) \le R_{\lambda}$), or if the point is dropped.
\begin{figure}[!ht]
\centering
\includegraphics[width=1\linewidth]{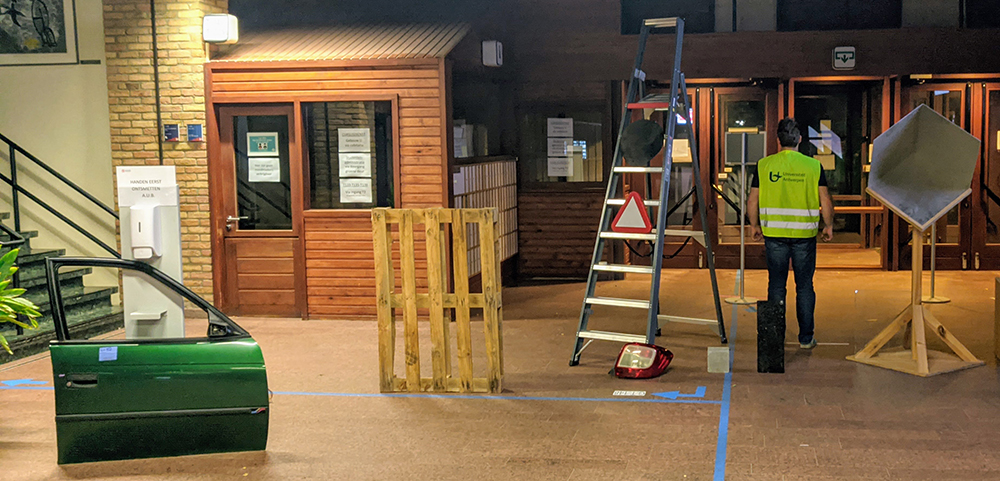}
\caption{Objects of various materials that were re-created in simulation to compare measurements of a real Ouster OS0-128 LiDAR sensor and its simulation counterpart at various distances to the objects.}
\label{fig:real_objects}
\end{figure}
\section{Experimental Validation}
\label{sec:results}

For validation of the proposed simulation models, we check if the simulation model approaches a real sensor to a sufficient level for various use-cases such as application testing, transfer-learning and others. We compare real-world recordings with an Ouster OS0-128 LiDAR (128 beam channels, $FOV_v$ of \ang{90}, 1024 samples over a $FOV_h$ of 360°) \cite{OS0Lidar} of several objects made of different materials which can be seen in Fig. \ref{fig:real_objects}. Subsequently, we re-created an accurate analogue of the real scene in the simulation framework.
We used manual measurements and an Intel Realsense D435i depth camera, in combination with the Open3D software library\cite{ZhouOpen3D:Processing} for facilitating scene reconstruction of the more complex objects. We use the ECOSTRESS data as input for the various surface materials and their reflectance value at the wavelength of the Ouster OS0-128 which is at \SI{850}{\nm}. From its data sheet we can also extract the maximum range $d_{max}$ of \SI{50}{\m} at a reflectance value $R_{L,max}$ of 80\%. In a first validation we compare measurements at different ranges to the objects and calculate the error in both the depth and intensity values for each beam of the LiDAR sensor between real and simulated points. 
\begin{table}[!ht]
\centering
\caption{Intensity and Depth Results by Distance to Objects}
\label{table:intensity_depth_distance}
\resizebox{0.7\columnwidth}{!}{%
\begin{tabular}{@{}c|cc|cc@{}}
Distance & \multicolumn{2}{c|}{Depth Error} & \multicolumn{2}{c}{Intensity Error} \\ \cline{2-5}
\addlinespace[0.1em]
to Objects & Mean & Std Dev & Mean & Std Dev \\ \hline
\addlinespace[0.1em]
\SI{33}{\m}  & \SI{14.65}{\cm} & \SI{12.69}{\cm} & 10.65 & 12.60 \\ 
\SI{25}{\m}  & \SI{13.54}{\cm} & \SI{11.63}{\cm} & 8.45 & 12.33 \\ 
\SI{18}{\m}  & \SI{10.69}{\cm} & \SI{10.15}{\cm} & 10.69 & 11.88 \\ 
\SI{12}{\m}  & \SI{8.65}{\cm} & \SI{9.77}{\cm} & 8.65 & 11.68 \\ 
\SI{7}{\m}  & \SI{6.41}{\cm} & \SI{8.13}{\cm} & 7.11 & 10.23 \\ 
\SI{5}{\m}  & \SI{6.25}{\cm} & \SI{6.80}{\cm} & 8.46 & 11.95 \\ 
\SI{4}{\m}  & \SI{5.88}{\cm} & \SI{5.46}{\cm} & 10.48 & 12.38 \\ 
\SI{2,5}{\m}  & \SI{5.12}{\cm} & \SI{4.74}{\cm} & 11.35 & 12.60 \\ 
\end{tabular}%
}
\end{table}
\begin{table}[!ht]
\centering
\caption{Intensity and Depth Results by Surface Material}
\label{table:intensity_depth_material}
\resizebox{0.7\columnwidth}{!}{%
\begin{tabular}{@{}c|cc|cc@{}}
Surface & \multicolumn{2}{c|}{Depth Error} & \multicolumn{2}{c}{Intensity Error} \\ \cline{2-5}
\addlinespace[0.1em]
Material & Mean & Std Dev & Mean & Std Dev \\ \hline
\addlinespace[0.1em]
Car Paint  & \SI{6.14}{\cm} & \SI{8.49}{\cm} & 9.11 & 11.68 \\ 
Car Light & \SI{10.43}{\cm} & \SI{13.41}{\cm} & 6.66 & 13.90 \\ 
Reflector  & \SI{13.54}{\cm} & \SI{13.45}{\cm} & 20.48 & 24.82 \\ 
Rubber  & \SI{8.44}{\cm} & \SI{11.98}{\cm} & 9.48 & 13.41 \\ 
Concrete  & \SI{9.00}{\cm} & \SI{7.65}{\cm} & 9.30 & 12.16 \\ 
Asphalt & \SI{8.68}{\cm} & \SI{3.15}{\cm} & 8.99 & 14.69 \\ 
Glass  & \SI{0.00}{\cm} & \SI{0.00}{\cm} & 0.00 & 0.00 \\ 
Wood & \SI{5.45}{\cm} & \SI{4.79}{\cm} & 4.19 & 2.68 \\ 
Aluminium & \SI{18.41}{\cm} & \SI{15.12}{\cm} & 17.10 & 14.30 \\ 
\end{tabular}%
}
\end{table}
\newpage 
\noindent These results are shown in Table \ref{table:intensity_depth_distance} and show that the further away from the objects, the error becomes larger. The accuracy and precision overall is acceptable and perceived as similar with that of the recreated objects using the reconstruction methods. Similarly, in Table \ref{table:intensity_depth_material} we show the analysis of the depth and intensity error but now comparing it between the various surface materials. Here a clear difference can be observed between the materials. As the ECOSTRESS data is interpolated for all incidence angles in our model, as well as being not a perfect match for the materials used in the experiments, these differences are expected. Furthermore, the lack of more incidence angle data points in the library is shown by the lower precision of some of the results. Similar limitations of using the ECOSTRESS library in this context were also noted by Muchkenhuber et al. \cite{Muckenhuber2020AutomotiveCapabilities}.
Finally, the real-time simulation using GPU-acceleration proved to keep up with the general simulation rate by the AirSim framework.

\section{Conclusions \& Future Works}
\label{sec:conclusion}
The simulation of the LiDAR sensor provides the necessary simulation of depth, intensity and instance segmentation labels to serve a variety of use-cases. The experimental results show decent enough accuracy and precision. However, some clear shortcomings are evident as the ECOSTRESS data is too limited in scope and detail. The library is too restrictive in the man-made material section for developing complex and varied environments that are typical for typical LiDAR applications. A much broader and larger dataset is needed for more man-made materials and at various incidence angles for several common LIDAR wavelengths. Finally, several other LiDAR physical aspects such as weather influence, beam divergence and multi-echo LIDAR can also be considered for future iterations of the simulation. 

\newpage
\bibliographystyle{IEEEtran}
\bibliography{main.bbl}

\end{document}